\documentclass[letterpaper, 10 pt, conference]{ieeeconf}  

\IEEEoverridecommandlockouts                              

\usepackage{amsthm}
\usepackage{amsmath}
\usepackage{amsfonts}
\usepackage{graphicx}
\usepackage{cuted}
\usepackage{bbm}

\usepackage{multirow} 
\usepackage{booktabs}
\usepackage{caption}
\usepackage{subcaption}
\usepackage{breqn}
\usepackage{dsfont}
\let\mathbbm\mathds

\newtheorem{theorem}{Theorem}
\newtheorem{lemma}[theorem]{Lemma}
\def\thickhline{\noalign{\hrule height1.2pt}}

\title{\LARGE \bf
Controllable Motion Generation via Diffusion Modal Coupling
}

\author{
Luobin Wang$^{*,1}$, Hongzhan Yu$^{*,1}$, Chenning Yu$^1$, Sicun Gao$^1$, Henrik Christensen$^1$ 
\thanks{$^*$Equal Contribution, $^{1}$University of California, San Diego}
}

\begin{document}

\maketitle
\thispagestyle{empty}
\pagestyle{empty}

\maketitle
\begin{abstract}
Diffusion models are increasingly used in robotics to represent multi-modal distributions over system states and behaviors,
but precise control of generated outcomes without degrading physical realism remains challenging.
This paper introduces a controllable diffusion framework that
(i) replaces the standard unimodal Gaussian prior with an explicit multi-modal prior, 
and (ii) enforces modal coupling between prior components and principal data modes through novel forward and reverse diffusion processes.
Sampling is initialized directly from a selected prior mode aligned with task constraints, 
avoiding train–test mismatch and manifold drift commonly induced by post‑hoc guidance.
Empirical evaluations on motion prediction (Waymo Dataset) and multi‑task control (Maze2D) show consistent improvements over guidance‑based baselines in fidelity, diversity, and controllability. 
These results indicate that multi‑modal priors with strong modal coupling provide a scalable basis for controllable motion generation in robotics.
\end{abstract}    
\section{INTRODUCTION}

Diffusion models~\cite{ho2020denoising} provide expressive, high-fidelity generative capabilities well suited to robotics,
where uncertainty and multi-modality are intrinsic.
Recent applications span sensor simulation~\cite{ran2024towards, zhou2024text2pde},
trajectory generation~\cite{jiang2023motiondiffuser, yang2024diffusion},
and control policy learning~\cite{chi2023diffusion, zhou2024diffusion},
illustrating the utility of diffusion for modeling complex, high-dimensional behavior distributions.
Despite this progress,
controllability, i.e., the ability to produce samples that satisfy task objectives and domain constraints while remaining on the data manifold,
remains a central obstacle.
Unconstrained sampling can produce implausible plans or policies, thereby undermining the intended driving objectives \cite{jiang2023motiondiffuser}.
Due to limited controllability, 
practitioners resort to generating large numbers of samples to capture desirable outcomes,
which raises scalability concerns and complicates the mining of high-fidelity samples for large-scale synthetic generations.



\begin{figure}[t]
    \centering
    \includegraphics[width=\columnwidth]{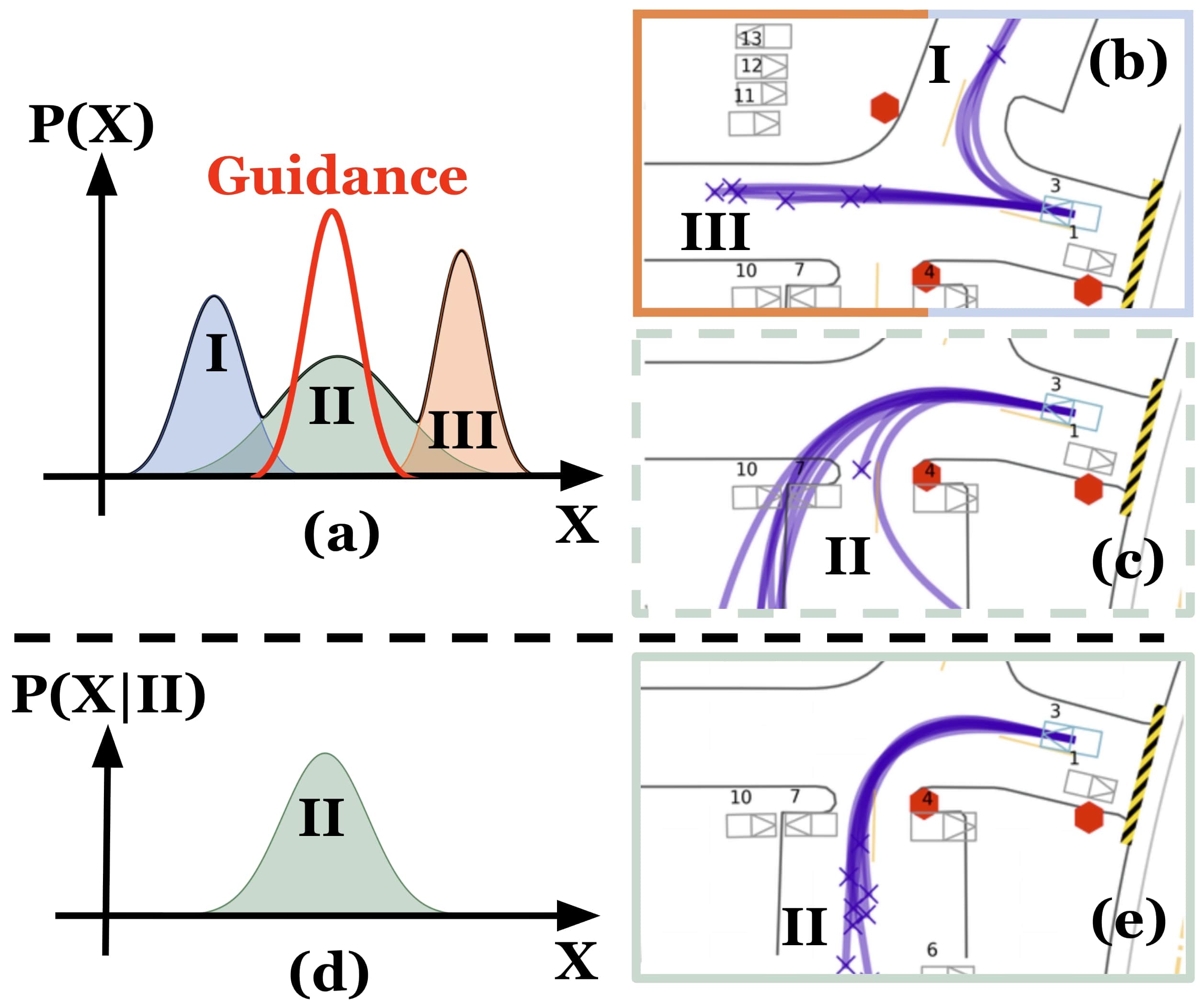}
    \caption{\small
    High-level comparison of guidance-based approaches versus our proposed method. 
    \textbf{(a)} A standard diffusion model fits a multi-modal data distribution (three modes).
    A guidance term (red) attempts to steer sampling toward a rare yet operationally critical mode (Mode II).
    \textbf{(b)} Standard (unguided) sampling concentrates on high-probability modes.
    \textbf{(c)} Guidance perturbs intermediate states off the well-trained data manifold,
    degrading fidelity.
    \textbf{(d)(e)} Our method couples each principal mode to a dedicated prior component,
    enabling direct, mode-aligned control at sampling while avoiding guidance-induced distribution mismatch.
    }
    \label{fig:guidance_vs_ours}
\end{figure}

Existing approaches typically inject constraints at inference,
via constraint-based sampling or post-hoc guidance,
to encode domain knowledge and task objectives into the reverse process~\cite{ho2022classifier, bansal2023universal},.
While such mechanisms improve alignment with target objectives, 
they introduce a train-test mismatch:
external gradients modify the learned score field during sampling,
perturbing the reverse dynamics and steering intermidiate states away from high‑density regions of the data distribution.
The resulting off-manifold drift degrades sample fidelity.
Achieving fine-grained controllability without sacrificing data-manifold realism therefore remains an open problem (Figure \ref{fig:guidance_vs_ours} [a-c]).

\begin{figure*}[!t]
    \centering
    \includegraphics[width=\textwidth]{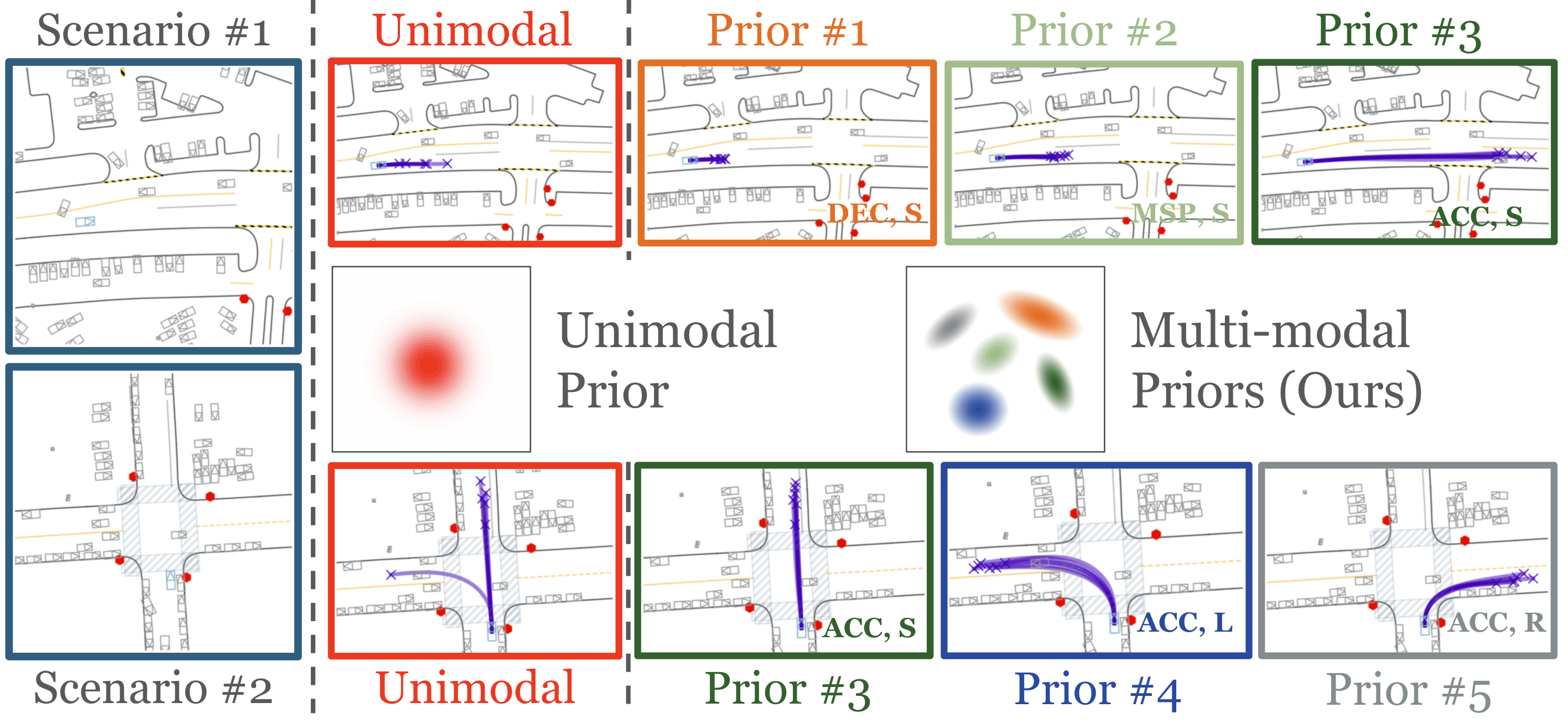}
    \caption{\small
    High-level overview.
    Conventional diffusion models use a unimodal prior distribution and lack an intrinsic mechanism to select which trajectories are emphasized.
    We introduce a multi-modal prior and enforce strong modal coupling between prior and data via a novel diffusion process. 
    The framework enables direct mode selection even with an unconditioned diffusion model,
    supporting precise and adaptive motion generation.
    In the figure, each prior component corresponds to one behavior. ``ACC'', ``DEC'' and ``MSP'' refer to speed modes (acceleration, deceleration, and maintaining speed), while 
    ``R'', ``L'' and ``S'' represent steering modes (right, left, and straight).
    }
    \label{fig:overview}
\end{figure*}

We address this problem by introducing modal coupling via a multi-modal prior.
Concretely, we replace the standard Gaussian prior with a Gaussian-mixture prior whose components are in one-to-one correspondence with principal modes of the data distribution (Figure \ref{fig:guidance_vs_ours} [d-e]).
We derive modified forward and reverse diffusion processes that (i) map all data samples sharing the same label to a common, non-standard Gaussian during noising and (ii) recover the corresponding data mode when denoising is initialized from the associated prior component.
At inference, controllability is obtained by selecting the prior mode consistent with the target constraints and running standard denoising, without post-hoc guidance.
By removing external guidance terms, our method eliminates train–test mismatch and mitigates intermediate-step distribution drift, while retaining flexible controllability (Fig.~\ref{fig:overview}).


Our current formulation assumes that the data modes subject to control are explicitly known, 
yet the diffusion model itself does not require conditioning on these mode labels.
While this may appear restrictive, 
it provides a foundation for strong controllability over data with unknown key modes,
wherein the central challenge shifts to accurately identifying the appropriate prior mode from target constraints.
Nevertheless, by adopting a \textit{multi-modal} prior distribution, strong \textit{modal coupling}, and a carefully-designed \textit{ prior parametrization}, 
our method significantly outperforms guidance-based techniques
in both fidelity and controllability.
We validate these claims on the Waymo dataset for motion prediction, and in Maze2D for multi-task control.
The paper is organized as follows.
Section \ref{section:related} and \ref{section:preliminary} review related work and background.
We detail the proposed method in Section \ref{section:theory}.
Section \ref{section:experiment} presents the experimental results,
and Section \ref{section:conclusion} concludes.
\section{Related Work}
\label{section:related}
The multi-modal nature of human behaviors poses a great challenge for predicting realistic trajectories and control sequences. Diffusion models have proven effective in capturing this multi-modality within driving scenarios while closely adhering to real-world behavior distributions. SceneDM \cite{guo2023scenedm} utilizes a diffusion-based framework to model joint-distributions of all agents in a scene. SceneDiffuser \cite{pronovost2023scenario} employs a latent diffusion architecture derived from Bird's Eye View representations, whereas MotionDiffuser \cite{jiang2023motiondiffuser} demonstrates its capabilities of predicting realistic future trajectories that align with 
true data distribution
via PCA-compressed trajectory representations. 
Additionally, VBD \cite{huang2024versatile} jointly optimizes a motion predictor and a denoiser that share the same scene encoder, which further improves realism and versatility. However, achieving such realism and broad distribution coverage often requires drawing many random samples from a standard Gaussian prior that is unimodal.
Our approach enhances the realism of generated trajectories 
by incorporating a multi-modal prior that more effectively captures distinct data modes.

The typical strategy to control diffusion-based generation is incorporating domain-specific objectives into the generation process.
Classifier guidance \cite{dhariwal2021diffusion} guides the diffusion model with a separate cost function that encodes the objectives during sampling. 
Recent works \cite{jiang2023motiondiffuser, ctg, yang2024diffusiones, zheng2025diffusionplanner} introduce either language-model–generated or analytical guidance functions to pursue realism or safety-critical objectives.

On the other hand, Classifier-free guidance \cite{ho2022classifier} additionally optimizes a time-dependent conditional model to obtain guidance. This is widely adopted in other fields including text-to-image \cite{rombach2021latentdiff} and 3D objection \cite{Liu_2023_ICCV} generation.
However, the constraints imposed by guidance often degrade the realism of generated motion \cite{ctg} due to distribution mismatch between denoising steps and corresponding diffusion forward steps. 
Our approach avoids this issue by coupling prior and data modes and constructing a shared probability path for both forward and reverse diffusion.
\section{Background}
\label{section:preliminary}

Diffusion models are probabilistic generative models that synthesize new data by iteratively denoising an initial noise sample.
They first define a forward stochastic process that progressively perturbs real data $x_0$ into pure noise: 
\begin{align}
    q(x_{1:T} | x_0) &= \prod_{t=1}^{T} q(x_{t} | x_{t-1}), \\ 
    q(x_{t} | x_{t-1}) &= \sqrt{1 - \beta_{t}} x_{t-1} + \sqrt{ \beta_{t}} \epsilon_{t},
\end{align}
where $x_{1:T}$ denotes the sequence of noised samples, $\epsilon_{t}$ is standard Gaussian noise, and $\beta_{t}$ is forward variance.
Let $\alpha_{t} := 1 - \beta_{t}$ and $\bar{\alpha}_{t} := \prod_{i=1}^{t} \alpha_{i}$.
The above forward process yields the closed-form marginal:
\begin{align}
    q(x_{t} | x_{0}) &= \mathcal{N}(x_{t}; \sqrt{\bar{\alpha}_{t}}x_{0}, (1 - \bar{\alpha}_{t})I). \label{ddpm_forward} 
\end{align}
Choosing a variance schedule $\beta_{1:T}$ with $\bar{\alpha}_T \to 0$ ensures the prior $x_{T} \sim \mathcal{N}(0, I)$, i.e., a standard Gaussian distribution.


The reverse process removes noise step by step via Gaussian transitions.
Concretely,
\begin{align}
    p_{\theta}(x_{t-1} | x_{t}) &= \mathcal{N}(x_{t-1}; \mu_{\theta}(x_{t}), \sigma_{t}^{2}I), \\ 
    \sigma_{t}^{2} &= \frac{1 - \bar{\alpha}_{t-1}}{1 - \bar{\alpha}_{t}}\beta_{t},
\end{align}
where $\theta$ are the denoiser parameters trained by maximizing the evidence lower bound \cite{ho2020denoising}: 
$\max_{\theta}$  $-\log p_{\theta}(x_{0}|x_{1})$ $+$ $\sum_{t} D_{KL}[q(x_{t-1}|x_{t}, x_{0}) || p_{\theta}(x_{t-1}|x_{t})]$.
A common reparameterization predicts noise:
\begin{align}
\mu_{\theta}(x_{t}) = \frac{1}{\sqrt{\alpha_{t}}} (x_{t} - \frac{1 - \alpha_{t}}{\sqrt{1 - \bar{\alpha}_{t}}}\epsilon_{\theta}(x_{t}, t)),
\end{align}
where $\epsilon_{\theta}$ approximates the score $\nabla_{x_{t}}\log q(x_{t})$ across noise scales.
Alternatively, 
predicting the clean sample $x_{\theta}^{0}$ gives the closed-form posterior mean:
\begin{align}
    \mu_{\theta}(x_{t}) = \frac{\sqrt{\alpha_{t}}(1 - \bar{\alpha}_{t-1})}{1 - \bar{\alpha}_{t}} x_{t} + \frac{\sqrt{\bar{\alpha}_{t-1}}\beta_{t}}{1 - \bar{\alpha}_{t}}x_{\theta}^{0}(x_{t}, t). \label{ddpm_posterior}
\end{align}

\textit{Controllability} denotes the ability to enforce the specified constraints on generated samples.
A common approach is guidance~\cite{ho2022classifier} 
which augments the learned score at each reverse step with a task-specific cost $f$:
\begin{align}
    \hat{\epsilon}_{\theta}(x_{t}, t) = \epsilon_{\theta}(x_{t}, t) + \omega \nabla_{x_{t}}f(x_{t}), \label{eq:post_hoc}
\end{align}
where $\omega$ controls the strength of the guidance.
Here, $f$ is flexible:
it can encode hand-designed differentiable penalties (e.g., speed limits, action bounds, or label/style preferences) or be realized implicitly via a conditional score model, as in the popular classifier-free guidance approaches~\cite{ho2022cfg}.

\section{Enhanced Diffusion Controllability via Modal Coupling}
\label{section:theory}




A common strategy for controlling diffusion‑based generation is guidance. 
However, guidance is absent from diffusion model training and is applied only at sampling as a post-hoc modification (\ref{eq:post_hoc}). 
This train–test mismatch can induce distribution shift, 
pushing intermediate states off the high‑fidelity data manifold.
To address this, 
we propose adopting a multi-modal prior and pose the question: 
\begin{center}
\begin{minipage}{0.9\columnwidth} 
\raggedright
\itshape
if each prior mode is tightly coupled to a principal data mode at training, 
can we run denoising process from different prior modes for direct controllability without guidance?
\end{minipage}
\end{center}
We show the answer is yes:
by modifying the diffusion processes to accommodate a carefully-parameterized Gaussian-mixture prior,
we can initialize the denoising/reverse process from a selected prior mode to target specific data constraints.
Importantly, 
this eliminates reliance on post-hoc modifications,
avoiding guidance-induced distribution mismatch.




\begin{figure}[t]
    \centering
    \includegraphics[width=\columnwidth]{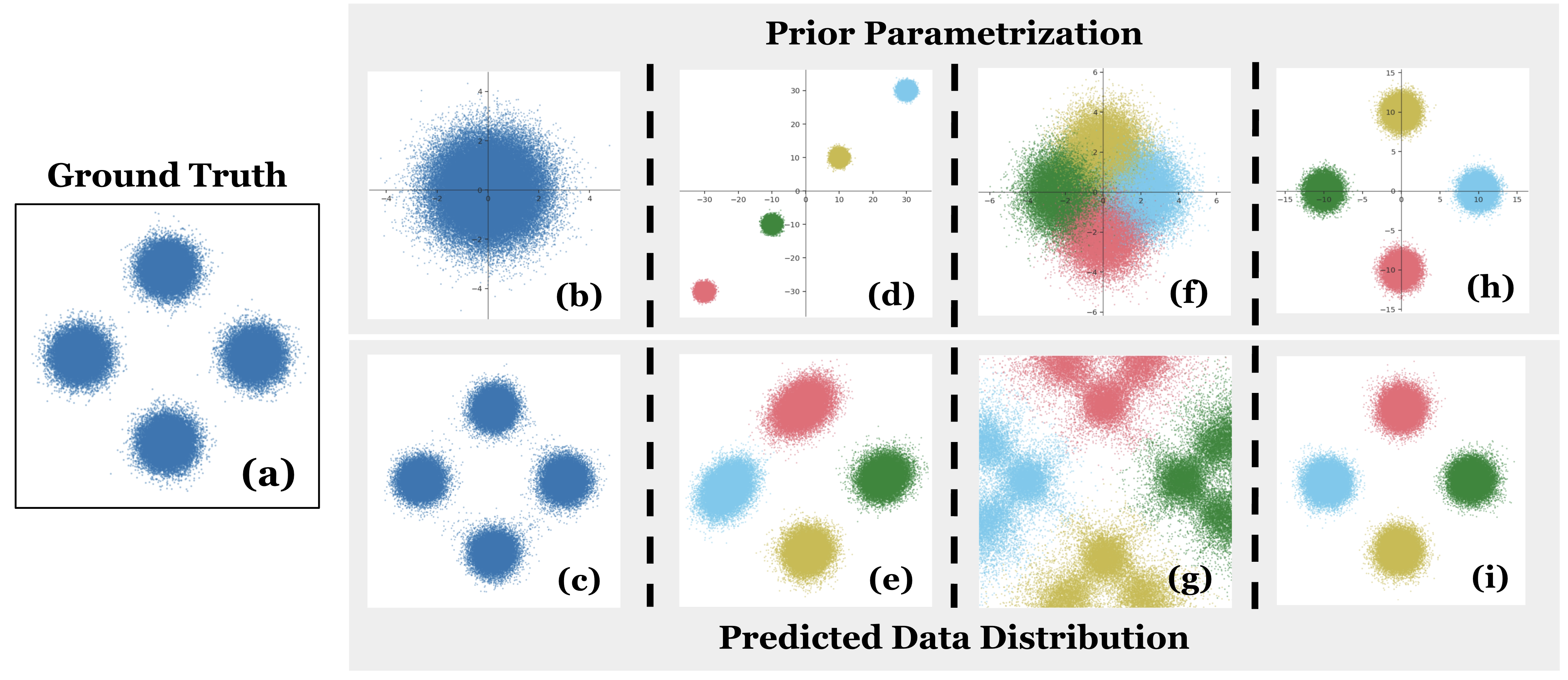}
    \caption{\small
    2D toy example.
    (a) The data distribution with four distinct modes.
    (b-c) Results from DDPM~\cite{ho2020denoising} show using unimodal prior yields spurious samples in the gaps between modes.
    (d-e) When prior means have large magnitude (i.e., lie far from the origin), the diffusion model struggles to recover realistic per-mode data distributions.
    (f-g) Insufficient separation between prior modes also  prevents the model from accurately capturing the data distribution.
    (h-i) With a carefully designed prior parameterization that maintains clear separation between modes without introducing excessive values,
    our method produces substantially fewer spurious samples and further enables direct control over individual modes. 
    Corresponding modes and samples share the same color.
    }
    \label{fig:toy}
\end{figure}
    
\subsection{Gaussian Mixture as a Multi-Model Prior}
We assume data modes are explicitly known, and
model the multi-modal prior as a Gaussian mixture model:
\begin{align}
    x_{T} \sim \sum_{i=1}^{k} r_{i} \cdot \mathcal{N}(\mu_{i}, \sigma_{i}^{2}I), \label{multimodal_prior}
\end{align}
where $k$ is the number of modes, $r_{i}$ is the proportion of data with mode label $i$, and $\mu_{i}$, $\sigma_{i}^{2}$ are the mean and variance of the component $i$.
Conditioned on a specific label $L$, the prior reduces to a unimodal Gaussian:
\begin{align}
    x_{T} | x_{0} \sim \mathcal{N}(\mu_{L}, \sigma_{L}^{2}I), \label{unimodal_prior}
\end{align}
allowing us to explicitly account for different data modes while retaining a unimodal form given a specific label.

\subsection{Modal Coupling} 

Effective training with a Gaussian mixture prior requires two conditions.
First \textbf{(modal coupling)}: 
each prior mode must correspond one-to-one with a data mode. 
For label $L$, 
the forward process maps data with label $L$ to $\mathcal{N}(\mu_{L}, \sigma_{L}^{2}I)$,
while initializing the reverse process from that specific prior mode
recovers the data of label $L$.
Second \textbf{(trajectory separation)}:
throughout denoising, 
diffusion paths from distinct prior modes must remain sufficiently separated to
preserve mode identity and avoid cross-mode interference.
This section focuses on modal coupling, 
while the discussion of prior parameterization for trajectory separation is in Section \ref{subsection:prior}.

We first define a forward noising process such that the terminal distribution satisfies a general Gaussian prior $x_{T}|x_{0} \sim \mathcal{N}(\mu, \sigma^{2}I)$.
\begin{lemma}\label{lemma1}
    Let $\eta_{t} := 1 + \sum_{m = 1}^{t-1} \Big(\sqrt{\prod_{n = m + 1}^{t}\alpha_{n}}\Big)$,
    and consider the forward noising process
    \begin{align}
        q(x_{t}|x_{t-1}) = \sqrt{\alpha_{t}}x_{t-1} + \sqrt{1 - \alpha_{t}}\sigma\epsilon_{t} + \frac{\mu}{\eta_{T}}.  \label{forward}
    \end{align}
    where $\epsilon_{t} \sim \mathcal{N}(0, I)$.
    Then, for any step $t$,
    \begin{align}
        q(x_{t}|x_{0}) = \mathcal{N}(x_{t} | \sqrt{\bar{\alpha}_{t}}x_{0} + \frac{\eta_{t}\mu}{\eta_{T}}, (1 - \bar{\alpha}_{t})\sigma^{2}I). \label{posterior}
    \end{align}
    Under the standard assumption that $\bar{\alpha}_{T} \to 0$ as $T$ grows large, it follows that $q(x_{T}|x_{0}) = \mathcal{N}(x_{T}; \mu, \sigma^{2}I)$.
\end{lemma}
\noindent\textit{Proof.} See Appendix \ref{appendix:lemma1}.
It shows that a \textit{constant} shift term $\mu/\eta_{T}$ at each forward step and noise scaling by $\sigma$ yield the desired terminal prior, enabling a matched reverse process.
\begin{lemma} \label{lemma2}
    For the diffusion model with the forward process defined in (\ref{forward}), the reverse process is:
    \begin{align}
        p(x_{t-1}|x_{t}) &= \mathcal{N}(x_{t-1} | \mu(x_{t}), \beta(x_{t})), \label{reverse}
    \end{align}
    where $\beta(x_{t}) = \frac{1 - \bar{\alpha}_{t-1}}{1 - \bar{\alpha}_{t}}\beta_{t} \sigma^{2}$, and
    \begin{align}
        \mu(x_{t}) &= \frac{\sqrt{\alpha_{t}}(1 - \bar{\alpha}_{t-1})}{1 - \bar{\alpha}_{t}} x_{t} + \frac{\sqrt{\alpha_{t-1}}\beta_{t}}{1 - \bar{\alpha}_{t}} \underline{\mathbf{x_{0}}} \label{one_step_reverse} \\ 
        &\hspace{0.4cm}+ \frac{\eta_{t-1}(1 - \alpha_{t}) - \sqrt{\alpha_{t}}(1 - \bar{\alpha}_{t-1})}{(1 - \bar{\alpha}_{t})\eta_{T}} \cdot \mu, \nonumber\\
        &= \frac{1}{\sqrt{\alpha_{t}}}(x_{t} - \frac{\mu}{\eta_{T}} - \frac{1 - \alpha_{t}}{\sqrt{1 - \bar{\alpha}_{t}}}\sigma \underline{\boldsymbol{\epsilon_t}}).
    \end{align}
\end{lemma}
\noindent\textit{Proof.} See Appendix \ref{appendix:lemma2}.
This completes the derivation of the modified forward and reverse processes, 
making each data mode directly associated with a non-standard Gaussian prior component. 
We now introduce the training objective.
Given a labeled dataset $\mathcal{X}$ and pre-defined prior parameters $\{\mu_{1:k}, \sigma_{1:k}\}$, 
we construct noisy samples via (\ref{posterior}) and train the clean-prediction model $x_{\theta}^{0}$ with:
\begin{align}
    \min_{\theta} \underset{\substack{t \in [1, T] \\\epsilon\sim\mathcal{N}(0, I)}}{\mathbb{E}} \Bigg[ \sum_{(x_{0}, L) \in \mathcal{X}} \| x_{\theta}^{0}(\hat{x}_{t}(x_{0}, L, \epsilon), t) - x_{0}\|^{2} \Bigg],\\
    \hat{x}_{t}(x_{0}, L, \epsilon) = \sqrt{\bar{\alpha}_{t}}x_{0} + \sqrt{1 - \bar{\alpha}_{t}}\sigma_{L} \epsilon + \frac{\eta_{t}\mu_{L}}{\eta_{T}}.
\end{align}
Switching the reparameterization of $\mu_{\theta}$ to noise prediction is straightforward but omitted for performance considerations.

At sampling, we can draw the prior sample $x_{T}$ from the mixture (\ref{multimodal_prior}) to cover all modes,
or fix a component (\ref{unimodal_prior}) to target specific constraints.

\subsection{Prior Parametrization}
\label{subsection:prior}

Ensuring trajectory separation is crucial for reliable mode identification,
which in turn underpins controllability. 
Without it,
reverse trajectories from different prior components may overlap, 
causing mode indistinguishability and loss of control (Figure \ref{fig:toy}). 
We address this by carefully parameterizing the Gaussian mixture prior.




To place the Gaussian means, we draw on the concept of placing \textit{evenly spaced} points on a high-dimensional sphere.
Let $d$ be the data dimension, and $k$ the number of modes.
If $k \leq d+1$, 
which is common in complex data distributions specialized by diffusion models, 
the means can be located at the vertices of a $(k-1)$-simplex embedded in $\mathbb{R}^{d}$~\cite{coxeter1973regular}.
Let $e_{1:k} \subset \mathbb{R}^{k}$ be the standard basis vectors and $\mathds{1}_{k}$ the all‑ones vector.
With sphere radius $\delta>0$, define
\begin{align}
    w_{i} &= \delta \cdot \sqrt{\frac{k}{k+1}} \cdot (e_{i} - \mathds{1}_{k} \cdot \frac{1}{k}) \\ 
    \mu_{i} &= [
        w_{i}^{1}, ..., w_{i}^{k}, 0, ..., 0]^{T} \in \mathbbm{R}^{d},
\end{align}
where $w_{i}^{j}$ denotes the $j$-th component of $w_{i}$.
One can verify that $\|\mu_{i}\|^{2} = \delta$ for all $i$, and the pairwise distance $d$ is:
\begin{align}
    d = \delta \cdot\sqrt{2+ 2 / (k-1)}.
\end{align}
Finally, we choose the Gaussian variances so that $c$-level confidence ellipsoids do not overlap.
That is, for all $i \in [1, k]$, 
\begin{align}
    \sigma_{i} \sqrt{\mathcal{X}^{2}_{d, c}} \leq d, \label{confidence_interval}
\end{align}
where $\mathcal{X}^{2}_{d, c}$ is the $c$ quantile of the chi‐square distribution with $d$ degress of freedom \cite{wilson1931distribution}.

\section{Experiments}
\label{section:experiment}

\begin{table*}[th]
\centering
\setlength{\tabcolsep}{5.pt} 
\renewcommand{\arraystretch}{1.5} 
\small
\begin{tabular}{ccc|c|c|c|c|c|c|c|c}
\toprule
\multicolumn{3}{c|}{Method} &\begin{tabular}[c]{@{}c@{}}minADE  \\ $(\downarrow)$\end{tabular} & \begin{tabular}[c]{@{}c@{}}minFDE \\ $(\downarrow)$\end{tabular} & \begin{tabular}[c]{@{}c@{}}Collision  \\ $(\downarrow)$\end{tabular} & \begin{tabular}[c]{@{}c@{}} OffRoad \\ $(\downarrow)$\end{tabular} & \begin{tabular}[c]{@{}c@{}} ACC[ST] \\ $(\uparrow)$\end{tabular} & \begin{tabular}[c]{@{}c@{}} ACC[SP] \\ $(\uparrow)$\end{tabular} & \begin{tabular}[c]{@{}c@{}} ACC \\ $(\uparrow)$\end{tabular} & \begin{tabular}[c]{@{}c@{}} Inference Time \\ $(\downarrow)$\end{tabular} \\ \thickhline 
\multicolumn{1}{c|}{\multirow{3}{*}{\textit{CG}}} 
& \multicolumn{2}{l|}{$\eta = 1$ \hspace{0.15cm}  } & 2.614 & 5.999  & 0.054 & 0.127 & 0.882 & 0.968 & 0.746 & 2.717 \\ \cline{2-11} \multicolumn{1}{c|}{}                      
& \multicolumn{2}{l|}{$\eta = 10$ \hspace{0.15cm} } & 2.862 & 6.655  & 0.057 & 0.160 & 0.916 & 0.987 & 0.902 & 2.703 \\ \cline{2-11} \multicolumn{1}{c|}{}                      
& \multicolumn{2}{l|}{$\eta = 100$} & 4.681 & 12.358 & 0.059 & 0.275 & \textbf{0.976} & 0.996 & \textbf{0.970} & 2.712 \\ \thickhline 
\multicolumn{1}{c|}{\multirow{4}{*}{\textit{CFG}}} 
& \multicolumn{2}{l|}{$\lambda = 0.9$ \hspace{0.15cm}}& 3.994 & 10.194 & 0.070 & 0.162 & 0.878 & 0.833 & 0.746 & 1.535 \\ \cline{2-11} \multicolumn{1}{c|}{}                      
& \multicolumn{2}{l|}{$\lambda = 1$ \hspace{0.15cm}}  & 3.034 & 7.138  & 0.069 & 0.159 & 0.877 & 0.977 & 0.864 & 1.566 \\ \cline{2-11} \multicolumn{1}{c|}{}                      
& \multicolumn{2}{l|}{$\lambda = 1.05$ \hspace{0.15cm}} & 2.687 & 6.055 & 0.065 & 0.157 & 0.885 & 0.996 & 0.883 & 1.544 \\  \cline{2-11} \multicolumn{1}{c|}{} 
& \multicolumn{2}{l|}{$\lambda = 1.1$}  & 3.099 & 7.506 & 0.062 & 0.184 & 0.887 & \textbf{1.000} & 0.887 & 1.547 \\ \thickhline 
\multicolumn{1}{c|}{\multirow{4}{*}{Ours}} 
& \multicolumn{2}{l|}{$\delta = 0.5$ \hspace{0.15cm} } & 2.506 & 5.104 & 0.053 & 0.121 & 0.895 & 0.999 & 0.894 & 0.899 \\ \cline{2-11} \multicolumn{1}{c|}{}                     
& \multicolumn{2}{l|}{$\delta = 1.0$ \hspace{0.15cm}}  & \textbf{2.266} & \textbf{4.757} & \textbf{0.043} & \textbf{0.092} & 0.914 & \textbf{1.000} & 0.914 & 0.892 \\ \cline{2-11} \multicolumn{1}{c|}{}   
& \multicolumn{2}{l|}{$\delta = 2.0$ \hspace{0.15cm}}  & 2.525 & 5.303 & 0.489 & 0.101 & 0.907 & 0.998 & 0.904 & 0.906 \\ \cline{2-11} \multicolumn{1}{c|}{}
& \multicolumn{2}{l|}{$\delta = 4.0$ \hspace{0.15cm}}  & 2.772 & 5.697 & 0.060 & 0.135 & 0.895 & 0.993 & 0.888 & \textbf{0.864} \\ \cline{2-11} \multicolumn{1}{c|}{}
& \multicolumn{2}{l|}{\textit{EM}}  & 2.578 & 5.500 & 0.060 & 0.140 & 0.897 & 0.995 & 0.893 & 0.891 \\              
\bottomrule
\end{tabular}
\caption{\small
    Quantitative results on the Waymo Open Motion Dataset under reference driving modes. Driving modes comprise steering and speed categories. minADE/minFDE are averaged across steering–speed modes, mitigating the effect of imbalanced mode frequencies. 
}
\label{table::waymo}
\vspace*{-10pt}
\end{table*}

\begin{figure*}[t]
    \centering
    \includegraphics[width=\textwidth]{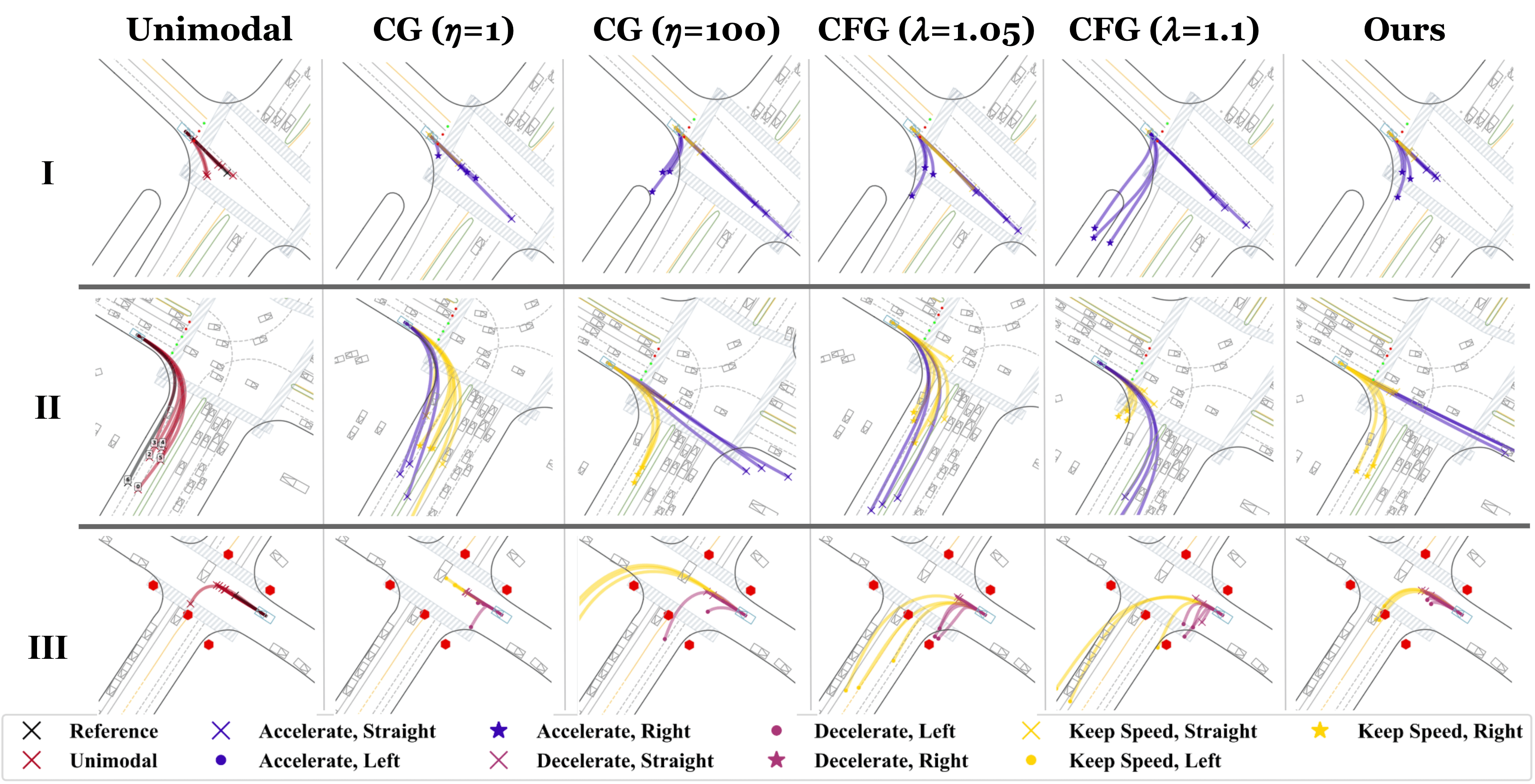}
    \caption{\small
Qualitative results for motion prediction. 
In standard diffusion, trajectories are sampled randomly from a unimodal prior, offering no inherent controllability.
CG applies guidance to steer generation, but it  relies heavily on the guidance influence factor, making it difficult to balance sample fidelity against controllability.
CFG blends unconditional and conditional outputs for guidance, which limits controllability and fidelity when the target lies off the reference data manifold.
Our method integrates modal coupling with a multi-modal prior distribution, yielding notable improvements in both sample fidelity and controllability.
    }
    \label{fig:waymo_control}
\vspace*{-10pt}
\end{figure*}

\begin{table}[t]
\centering
\setlength{\tabcolsep}{5.pt} 
\renewcommand{\arraystretch}{1.5} 
\small
\begin{tabular}{ccc|c|c|c}
\toprule
\multicolumn{3}{c|}{Method} & \begin{tabular}[c]{@{}c@{}} Collision \\ $(\downarrow)$\end{tabular} & \begin{tabular}[c]{@{}c@{}}OffRoad \\ 
$(\uparrow)$\end{tabular} & \begin{tabular}[c]{@{}c@{}} ACC \\ $(\uparrow)$\end{tabular} \\ \hline 
\multicolumn{1}{c|}{\multirow{3}{*}{\textit{CG}}} 
& \multicolumn{2}{l|}{$\eta = 1$ \hspace{0.15cm}  } & 0.050 & 0.125   & 0.595 \\ \cline{2-6} \multicolumn{1}{c|}{}                      
& \multicolumn{2}{l|}{$\eta = 10$ \hspace{0.15cm} } & 0.060 & 0.135   & 0.830 \\ \cline{2-6} \multicolumn{1}{c|}{}                                
& \multicolumn{2}{l|}{$\eta = 100$} & 0.060 & 0.205  & \textbf{0.960} \\ \thickhline 
\multicolumn{1}{c|}{\multirow{4}{*}{\textit{CFG}}} 
& \multicolumn{2}{l|}{$\lambda = 0.9$ \hspace{0.15cm}}& 0.080 & 0.215 & 0.725 \\ \cline{2-6} \multicolumn{1}{c|}{}                                
& \multicolumn{2}{l|}{$\lambda = 1$ \hspace{0.15cm}}  & 0.075 & 0.195 & 0.875 \\ \cline{2-6} \multicolumn{1}{c|}{}                                
& \multicolumn{2}{l|}{$\lambda = 1.05$ \hspace{0.15cm}} & 0.060 & 0.185 & 0.900 \\ \cline{2-6} \multicolumn{1}{c|}{}                                
& \multicolumn{2}{l|}{$\lambda = 1.1$}  & 0.060 & 0.195 & 0.910 \\ \thickhline 
\multicolumn{1}{c|}{\multirow{4}{*}{Ours}} 
& \multicolumn{2}{l|}{$\delta = 0.5$ \hspace{0.15cm} } & 0.060 & 0.115 & 0.915 \\ \cline{2-6} \multicolumn{1}{c|}{}                                
& \multicolumn{2}{l|}{$\delta = 1.0$ \hspace{0.15cm}}  & \textbf{0.040} & \textbf{0.110} & 0.930 \\ \cline{2-6} \multicolumn{1}{c|}{}                                
& \multicolumn{2}{l|}{$\delta = 2.0$ \hspace{0.15cm}}  & 0.055 & 0.120 & 0.920 \\ \cline{2-6} \multicolumn{1}{c|}{}                                
& \multicolumn{2}{l|}{$\delta = 4.0$ \hspace{0.15cm}}  & 0.070 & 0.140 & 0.925 \\ \cline{2-6} \multicolumn{1}{c|}{}                                
& \multicolumn{2}{l|}{\textit{EM}}  & 0.060 & 0.180 &  0.905  \\   
\bottomrule
\end{tabular}
\caption{\small
    Quantitative validation for trajectory synthesis controllability,
    using $8$ sampled trajectories per scenario.
    The test set consists of $200$ scenarios, each manually labeled with potentially feasible ego-agent futures that deviate from the dataset's original trajectories. 
}
\label{table::waymo_control}
\end{table}

We begin with the Waymo dataset, showing that our approach produces realistic and feasible motions while preserving controllability, in contrast to post-hoc guidance baselines. 
We then show that our method extends naturally to a multi-task setting by treating each task as a distinct mode. Unlike prior diffusion-based planners that train a separate model per task \cite{chi2023diffusionpolicy, janner2022diffuser}, a single model suffices for our approach while maintaining strong per-task performance.

\subsection{Waymo - Controllable Motion Prediction}

We evaluate on the Waymo Open Motion Dataset~\cite{ettinger2021large},
under single-agent setting with explicitly-defined mode structure.
Following \cite{tan2023lctgen}, ground-truth actions $L$ are deterministically labeled along two independent axes:
steering (left, right, straight) and speed (accelerate, decelerate, maintain speed) with a differentiable classifier $L=g(x)$.
The task is to generate an $8$-s ego control trajectory consistent with the queried action label while respecting map topology and contextual agent interactions. 
We predict $2$-D controls over $80$ future steps (dimension $160$),
yielding $k=9$ action modes from the Cartesian product of the two axes.


Versatile Behavior Diffusion (VBD)~\cite{huang2024versatile} is adopted as the model backbone.
From VBD, we reuse: 
(i) the query-centric Transformer encoder for scene-level interaction encoding,
and (ii) diffusion timesteps embeddings for in-context conditioning.
Our instantiation departs from VBD in two ways:
(1) we flatten the control sequence as a single vector representation~\cite{zheng2025diffusionplanner},
and (2) during denoising, 
we apply cross-attention from the ego control-sequence embeddings to the scene context to capture ego-environment interactions.

We compare against two guidance baselines.
\textbf{Classifier Guidance} (\textit{CG}) applies the deterministic classifiers (used for labeling) to yield gradients with respect to the unconditioned score network~\cite{ctg}. 
\textbf{Classifier-Free Guidance} (\textit{CFG}) trains additional conditional models with shared parameters~\cite{ho2022cfg}.
Both methods apply guidance in a post-hoc manner, 
with standard formulations given as follows:
\begin{align*}
\mathbf{CG}  \quad & \hat{\epsilon}_{\theta}(x_{t}, t) = \epsilon_{\theta}(x_{t}, t) + \eta \nabla_{x_{t}}g(\mu_\theta(x_t)), \\
\mathbf{CFG} \quad & \hat{\epsilon}_{\theta}(x_{t}, t) = \epsilon_{\theta}(x_{t}, t) + \lambda (\epsilon_{\theta}(x_{t}, t, L) - \epsilon_{\theta}(x_{t}, t)),
\end{align*}
Unless stated otherwise, 
we set per-mode scales $\sigma_{i}=1$ for $i\in\{1,\dots, k\}$.
We also evaluate an \textbf{empirical-means} (\textit{EM}) initialization that sets the multi-modal prior's means and standard deviations from training-set statistics.


Next, we describe the evaluation metrics used in our study.
We report \textit{Average Displacement Error (ADE)}, i.e., the mean Euclidean distance between predicted and ground-truth positions, and \textit{Final Displacement Error (FDE)}, i.e., the terminal position error. 
To assess physical plausibility, we compute \textit{Collision} and \textit{Off-Road} rates\cite{wosac}, defined as the fraction of rollouts that intersect another agent or leave the drivable area, respectively.
We introduce accuracy (\textit{ACC}) statistics to evaluate control fidelity, 
while further decomposing it into per-axis statistics: \textit{ACC[ST]} (steering) and \textit{ACC[SP]} (speed).

\subsubsection{Motion Prediction Accuracy}
Table \ref{table::waymo} summarizes the prediction accuracy results,
evaluated against ground-truth labels of the ego-agent’s future behaviors.
While the focus of this work is on controllability, 
we demonstrate that coupling diffusion model training on explicitly labeled datasets with a multi-modal prior distribution substantially improves predictive performance.

\textbf{Comparison to guidance baselines.}
For both guidance approaches (CG, CFG), the guidance coefficient ($\eta$ for CG; $\lambda$ for CFG) is sensitive to tuning and enforces a control-fidelity trade-off.
Since these methods operate with a unimodal prior,
they offer limited mode-specific learning and coverage.
Even at their best settings,
our method outperforms them:
a mode-coupled prior allocates capacity per mode and enables direct, label-aligned control,
yielding more robust predictions.
Computationally, 
CG requires per-step backpropagation through the diffusion model,
while CFG doubles forward passes.
Both introduce higher inference cost than our single reverse process from the selected prior component.

\textbf{Ablation on $\delta$ and EM prior.}
Setting $\delta=1.0$ consistently outperforms $\delta = 0.5$,
improving inter-component separation in the prior and reducing overlap among reverse-diffusion trajectories.
Excessively large $\delta$ degrades performance: 
as in the $2$-D toy example (Figure \ref{fig:toy}), 
components displaced far from the origin yield implausible trajectory distributions.
Under the EM configuration, 
steering means concentrate near zero because left/right-turn trajectories contain extended straight segments.
Speed statistics are less skewed but still exhibit partial mode collapse.
Although EM preserves cross-mode action-sequence correlations,
the increased component overlap reduces separability and harms accuracy.

\subsubsection{Controllable Trajectory Synthesis}

Table~\ref{table::waymo_control} evaluates realism and feasibility when generation is conditioned on driving modes that deviate from the dataset reference.
We manually annotate feasible future modes: 
e.g., if the ego is in a lane that permits turning while the reference continue straight, 
the turn is marked feasible.
The evaluation set contains $200$ scenarios and will be released with the code. 
Visualizations of generated trajectories using various controllability methods on manually labeled scenarios are presented in Figure \ref{fig:waymo_control}.

Under out-of-distribution conditioning,
our method continues to produce feasible, on-road trajectories, 
as reflected in the low collision and off-road rates.
CG/CFG often fail to realize off-reference modes with realistic generations: e.g., intersection scenes frequently go off-road after crossing.
While achieving superior sample realism, 
the proposed method also maintains stable, high accuracy.
These results underscore the limitations of guidance-based control (particularly post-hoc guidance) that our approach is designed to overcome.

\subsection{Maze2d - Multi-task Control}


We extend our method method to multi-task control.
Rather than confining control-level constraints to a single task,
we can also view each task itself as a distinct data mode.
We illustrate this with evaluations in Maze2D, 
where a single, task-agnostic model is trained to perform long-horizon path planning across three maze configurations.

\begin{table}[t]
\centering
\setlength{\tabcolsep}{4.5pt} 
\renewcommand{\arraystretch}{1.5} 
\small
\begin{tabular}{cc|c|c|c}
\toprule
\multicolumn{2}{c|}{Method}                             & U-Maze & Medium & Large \\ \hline
\multicolumn{1}{c|}{\multirow{3}{*}{Diffuser}} & U-Maze & 113.9  & N/A      & N/A     \\ \cline{2-5} 
\multicolumn{1}{c|}{}                          & Medium & N/A      & \textbf{121.5}  & N/A     \\ \cline{2-5} 
\multicolumn{1}{c|}{}                          & Large  & N/A      & N/A      & \textbf{123.0} \\ \hline
\multicolumn{2}{c|}{Ours, $k=3$, $\delta = 30$}                               & \textbf{119.5}  & 121.4  & 120.9 \\ 
\bottomrule
\end{tabular}
\caption{
\small Quantitative evaluations on Maze2D.
We adopt the baseline performance from \cite{janner2022planning}, using normalized accumulative reward returns as the evaluation metric.
Notably, the baseline trains a separate diffusion model for each layout (or mode), 
unlike in Table \ref{table::waymo} where our comparisons focus on multi-modal data modeling and the baseline there is a single model handling various modes. 
}
\label{table::maze2d}
\end{table}

We define the diffusion model to predict $384$-step state-control trajectories conditioned on the initial and goal positions. 
With $3$-D states and $2$-D controls,
the target distribution is $1920$-dimensional.
We set $\sigma_{i} = 1$ for $i \in [1, 3]$, 
and choose $\delta = 30$.
As shown in Table \ref{table::maze2d}, the unified model matches baseline performance across all tested mazes.
Note that the goal here is not to surpass the baseline,
but to demonstrate that casting tasks as modes establishes modal coupling over a multi-modal prior,
enabling diverse tasks to be handled by a single model without task-specific conditioning achitectures.


\section{Conclusion}
\label{section:conclusion}

This paper presents a novel framework that enables
fine-grained control over diffusion models while preserving high-fidelity sample generation.
By aligning the sampling process with key data modes from the outset,
our method avoids the distribution drift common in post-hoc guidance approaches.
Experimental evaluations show that the proposed method consistently outperforms existing techniques 
on both quantitative and qualitative measures.
Moreover, this work lays a strong foundation for future research aimed at relaxing the assumption of explicit known data modes, 
thereby advancing towards more controllable diffusion models.

\section*{APPENDIX}

\subsection{Proof of Lemma \ref{lemma1}}
\label{appendix:lemma1}

The proof for (\ref{posterior}) proceeds by induction.
We begin with the base case using the proposed forward process (\ref{forward}):
\begin{align}
    x_{1} &= \sqrt{\alpha_{1}} x_{0} + \sqrt{1 - \alpha_{1}} \sigma \epsilon_{1} + \frac{\mu}{\eta_{T}} \label{l1b1}\\
    &\sim \mathcal{N}(\sqrt{\bar{\alpha}_{1}} x_{0} +\frac{\eta_{1}\mu}{\eta_{T}}, (1 - \bar{\alpha}_{1})\sigma^{2}I) \label{l1b2}
\end{align}
The derivation from (\ref{l1b1}) to (\ref{l1b2}) is based on the fact that $\alpha_{1} = \bar{\alpha}_{1}$ and $\eta_{1} = 1$ by definition.
Next, we assume that for an arbitrary $t \in [2, T]$, it holds true that:
\begin{align}
    x_{t-1} = \sqrt{\bar{\alpha}_{t-1}} x_{0} + \sqrt{1 - \bar{\alpha}_{t-1}}\sigma\epsilon_{t-1} + \frac{\eta_{t-1}\mu}{\eta_{T}}.
\end{align}
From the forward process (\ref{forward}), we have:
\begin{align}
    x_{t} &= \sqrt{\alpha_{t}}x_{t-1} + \sqrt{1 - \alpha_{t}} \sigma \epsilon_{t} + \frac{\mu}{\eta_{T}} \\ 
    &= \sqrt{\alpha_{t}}\Big(\sqrt{\bar{\alpha}_{t-1}} x_{0} + \sqrt{1 - \bar{\alpha}_{t-1}}\sigma\epsilon_{t-1} + \frac{\eta_{t-1}\mu}{\eta_{T}}\Big) \\ 
    &\hspace{0.5cm} + \sqrt{1 - \alpha_{t}} \sigma \epsilon_{t} + \frac{\mu}{\eta_{T}} \nonumber \\ 
    &= \sqrt{\bar{\alpha}_{t}}x_{0} + \frac{\sqrt{\alpha_{t}}\eta_{t-1} + 1}{\eta_{T}} \cdot \mu \label{l1h1} \\ 
    &\hspace{0.5cm} + \Big(\sqrt{\alpha_{t}(1 - \bar{\alpha}_{t-1})}\sigma\epsilon_{t-1} + \sqrt{1 - \alpha_{t}} \sigma \epsilon_{t}\Big). \nonumber
\end{align}
Note that by the definition of $\eta_{t}$:
\begin{align}
    \eta_{t} &= 1 + \sum_{m=1}^{t-1}\Bigg(\sqrt{\prod_{n=m+1}^{t}\alpha_{n}}\Bigg) \\ 
    &= 1 + \sum_{m=1}^{t-1}\Bigg(\sqrt{\prod_{n=m+1}^{t-1}\alpha_{n}} \cdot \sqrt{\alpha_{t}}\Bigg) \\ 
    &= 1 + \sqrt{\alpha_{t}} \cdot \Big[1 + \sum_{m=1}^{t-2}\Bigg(\sqrt{\prod_{n=m+1}^{t-1}\alpha_{n}}\Bigg)\Big] \\ 
    &= 1 + \sqrt{\alpha_{t}}\eta_{t-1}.
\end{align}

Meanwhile, to handle the last term in (\ref{l1h1}),
we essentially merge two zero-mean Gaussian distributions with distinct variances.
That is, merging $\mathcal{N}(0, \sigma_{1}^{2}I)$ and $\mathcal{N}(0, \sigma_{2}^{2}I)$ leads to the new distribution $\mathcal{N}(0, (\sigma_{1}^{2} + \sigma_{2}^{2})I)$.
Here, the merged standard deviation is:
\begin{align}
    \alpha_{t}(1 - \bar{\alpha}_{t-1})\sigma^{2} + (1 - \alpha_{t})\sigma^{2} &= (1 - \bar{\alpha}_{t})\sigma^{2}.
\end{align}
Substituting everything back into (\ref{l1h1}), we have:
\begin{align}
    x_{t} &= \sqrt{\bar{\alpha}_{t}} x_{0} + \frac{\eta_{t}}{\eta_{T}}\mu + \sqrt{1 - \bar{\alpha}_{t}}\sigma\epsilon^{*}, \label{l1h2}
\end{align}
where $\epsilon^{*}$ denotes an arbitrary standard Gaussian sample.
This concludes the proof of Lemma \ref{lemma1}.

\subsection{Proof of Lemma \ref{lemma2}}
\label{appendix:lemma2}

First, the reverse probability is tractable only when conditioned on $x_{0}$.
By Bayes' theorem, we have:
\begin{align}
    p(x_{t-1} | x_{t}, x_{0}) &= q(x_{t} | x_{t-1}, x_{0}) \cdot \frac{q(x_{t-1} | x_{0})}{q(x_{t} | x_{0})}.
\end{align}
Then by Lemma \ref{lemma1}:
\begin{align}
    &\hspace{0.6cm}p(x_{t-1} | x_{t}, x_{0}) \\ 
    &\propto \exp (-\frac{1}{2} \Big[\frac{(x_{t} - \sqrt{\alpha_{t}} x_{t-1} - \mu/\eta_{T})^{2}}{(1 - \alpha_{t})\sigma^{2}} \label{l2l1}\\
    &\hspace{1cm}+ \frac{(x_{t-1} - \sqrt{\bar{\alpha}_{t-1}}x_{0} - \big(\eta_{t-1}/\eta_{T}\big)\mu)^{2}}{(1 - \bar{\alpha}_{t-1})\sigma^{2}} \nonumber \\ 
    &\hspace{1cm}- \frac{(x_{t} - \sqrt{\bar{\alpha}_{t}}x_{0} - \big(\eta_{t}/\eta_{T}\big)\mu)^{2}}{(1 - \bar{\alpha}_{t})\sigma^{2}}\Big]) \nonumber \\ 
    &\propto \exp (-\frac{1}{2} \Big[ \bigg( \frac{\alpha_{t}}{(1 - \alpha_{t})\sigma^{2}} + \frac{1}{(1 - \bar{\alpha}_{t-1}\sigma^{2})}\bigg) x_{t-1}^{2} \label{l2l2} \\ 
    &\hspace{2cm}+ 2\bigg(\frac{-\sqrt{\alpha_{t}}x_{t} + \sqrt{\alpha_{t}}\mu/\eta_{T}}{(1 - \alpha_{t})\sigma^{2}} \nonumber\\
    &\hspace{2cm}+ \frac{-\sqrt{\bar{\alpha}_{t-1}}x_{0} - (\eta_{t-1}/\eta_{T})\mu}{(1 - \bar{\alpha}_{t-1})\sigma^{2}} \bigg) x_{t-1}\Big]) \nonumber
\end{align}
From (\ref{l2l1}) to (\ref{l2l2}), 
the constant terms that do not involve $x_{t-1}$ are all omitted.
Following the standard Gaussian density function, 
the mean and variance of $p(x_{t-1} | x_{t}, x_{0})$ can be parameterized as $\mathcal{N}(\tilde{\mu}, \tilde{\beta})$ where:
\begin{align}
    \tilde{\beta} &= 1 / \bigg(\frac{\alpha_{t}}{(1 - \alpha_{t})\sigma^{2}} + \frac{1}{(1 - \bar{\alpha}_{t-1}\sigma^{2})}\bigg) \\ 
    &= \frac{1 - \bar{\alpha}_{t-1}}{1 - \bar{\alpha}_{t}} \cdot (1 - \alpha_{t}) \sigma^{2}.
\end{align}
Then we derive $\tilde{\mu}$ as follows:
\begin{align}
    \tilde{\mu}(x_{t}, x_{0}) &= -\bigg(\frac{-\sqrt{\alpha_{t}}x_{t} + \sqrt{\alpha_{t}}\mu/\eta_{T}}{(1 - \alpha_{t})\sigma^{2}} \\
    &\hspace{1cm}+ \frac{-\sqrt{\bar{\alpha}_{t-1}}x_{0} - (\eta_{t-1}/\eta_{T})\mu}{(1 - \bar{\alpha}_{t-1})\sigma^{2}} \bigg) \cdot \tilde{\beta}_{t} \nonumber\\
    &= \frac{\sqrt{\alpha_{t}}(1 - \bar{\alpha}_{t-1})}{1 - \bar{\alpha}_{t}} x_{t} + \frac{\sqrt{\bar{\alpha}_{t-1}}(1 - \alpha_{t})}{1 - \bar{\alpha}_{t}} x_{0} \\ 
    &\hspace{1cm}+ \frac{\eta_{t-1}(1 - \alpha_{t}) - \sqrt{\alpha_{t}}(1 - \bar{\alpha}_{t-1})}{(1 - \bar{\alpha}_{t}) \eta_{T}} \cdot \mu  \nonumber
\end{align}
Furthermore, we can parameterize $x_{0}$ in terms of $x_{t}$ and $\epsilon_{t}$ based on (\ref{l1h2}):
\begin{align}
    \tilde{\mu}_{t}(x_{t}) &= \frac{\sqrt{\alpha_{t}}(1 - \bar{\alpha}_{t-1})}{1 - \bar{\alpha}_{t}} x_{t} \\ 
    &+ \frac{\sqrt{\bar{\alpha}_{t-1}}(1 - \alpha_{t})}{1 - \bar{\alpha}_{t}} \cdot \frac{x_{t} - \sqrt{1 - \bar{\alpha}_{t}}\sigma\epsilon_{t} - (\eta_{t} / \eta_{T})\mu}{\sqrt{\bar{\alpha}_{t}}} \nonumber \\
    &+ \frac{\eta_{t-1}(1 - \alpha_{t}) - \sqrt{\alpha_{t}}(1 - \bar{\alpha}_{t-1})}{(1 - \bar{\alpha}_{t}) \eta_{T}} \cdot \mu \nonumber \\ 
    &= \frac{1}{\sqrt{\alpha_{t}}} \Big( x_{t} - \frac{1 - \alpha_{t}}{\sqrt{1 - \bar{\alpha}_{t}}}\sigma\epsilon_{t} \Big) \\
    &+ \frac{\mu}{(1 - \bar{\alpha}_{t})\eta_{T}} \cdot \Big( \eta_{t-1}(1 - \alpha_{t})\nonumber \\ 
    &\hspace{2cm}- \sqrt{\alpha_{t}}(1 - \bar{\alpha}_{t-1}) - \frac{(1 - \alpha_{t})\eta_{t}}{\sqrt{\alpha_{t}}}\Big)\nonumber 
\end{align}
\begin{align}
    &\hspace{-1.7cm}= \frac{1}{\sqrt{\alpha_{t}}} \Big( x_{t} - \frac{1 - \alpha_{t}}{\sqrt{1 - \bar{\alpha}_{t}}}\sigma\epsilon_{t} \Big) \\ 
    &\hspace{-0.7cm}+ \frac{\mu}{(1 - \bar{\alpha}_{t})\eta_{T}} \cdot \frac{-(1 - \bar{\alpha}_{t})}{\sqrt{\alpha_{t}}} \nonumber \\ 
    &\hspace{-1.7cm}= \frac{1}{\sqrt{\alpha_{t}}} \Big( x_{t} - \frac{\mu}{\eta_{T}} - \frac{1 - \alpha_{t}}{\sqrt{1 - \bar{\alpha}_{t}}}\sigma\epsilon_{t} \Big)
\end{align}
Last three equations are due to $\eta_{t} = 1 + \sqrt{\alpha_{t}} \eta_{t-1}$.

\bibliographystyle{unsrt}
\bibliography{reference}

\end{document}